\documentclass[runningheads]{llncs}
\usepackage[T1]{fontenc}
\usepackage{amssymb}
\usepackage{algorithm}  
\usepackage{algpseudocode}  
\usepackage{multirow}
\usepackage{subfigure}
\usepackage{graphicx}
\usepackage{amsmath}
\usepackage{hyperref}
\usepackage{booktabs}

\begin{document}
\title{A Visual Self-attention Mechanism Facial Expression Recognition Network beyond Convnext}
\titlerunning{A Visual Facial Expression Signal Feature Processing Network}
%
\author{Bingyu Nan
\and
Feng Liu\thanks{corresponding author.}
\and
Xuezhong Qian\thanks{corresponding author.}
\and
Wei Song}

\authorrunning{Bingyu Nan et al.}

\institute{Jiangnan University, Wuxi 214122, China\\
Shanghai Jiao Tong University, Shanghai 200062, China\\
\email{lsttoy@163.com, xzqian@jiangnan.edu.cn}}
\maketitle
\begin{abstract}
Facial expression recognition is an important research direction in the field of artificial intelligence. Although new breakthroughs have been made in recent years, the uneven distribution of datasets and the similarity between different categories of facial expressions, as well as the differences within the same category among different subjects, remain challenges. This paper proposes a visual facial expression signal feature processing network based on truncated ConvNeXt approach(Conv-cut), to improve the accuracy of FER under challenging conditions. The network uses a truncated ConvNeXt-Base as the feature extractor, and then we designed a Detail Extraction Block to extract detailed features, and introduced a Self-Attention mechanism to enable the network to learn the extracted features more effectively. To evaluate the proposed Conv-cut approach, we conducted experiments on the RAF-DB and FERPlus datasets, and the results show that our model has achieved state-of-the-art performance. Our code could be accessed at Github.
\end{abstract}

\section{Introduction}

\begin{figure}[!ht]
\centering
\includegraphics[width=0.7\textwidth]{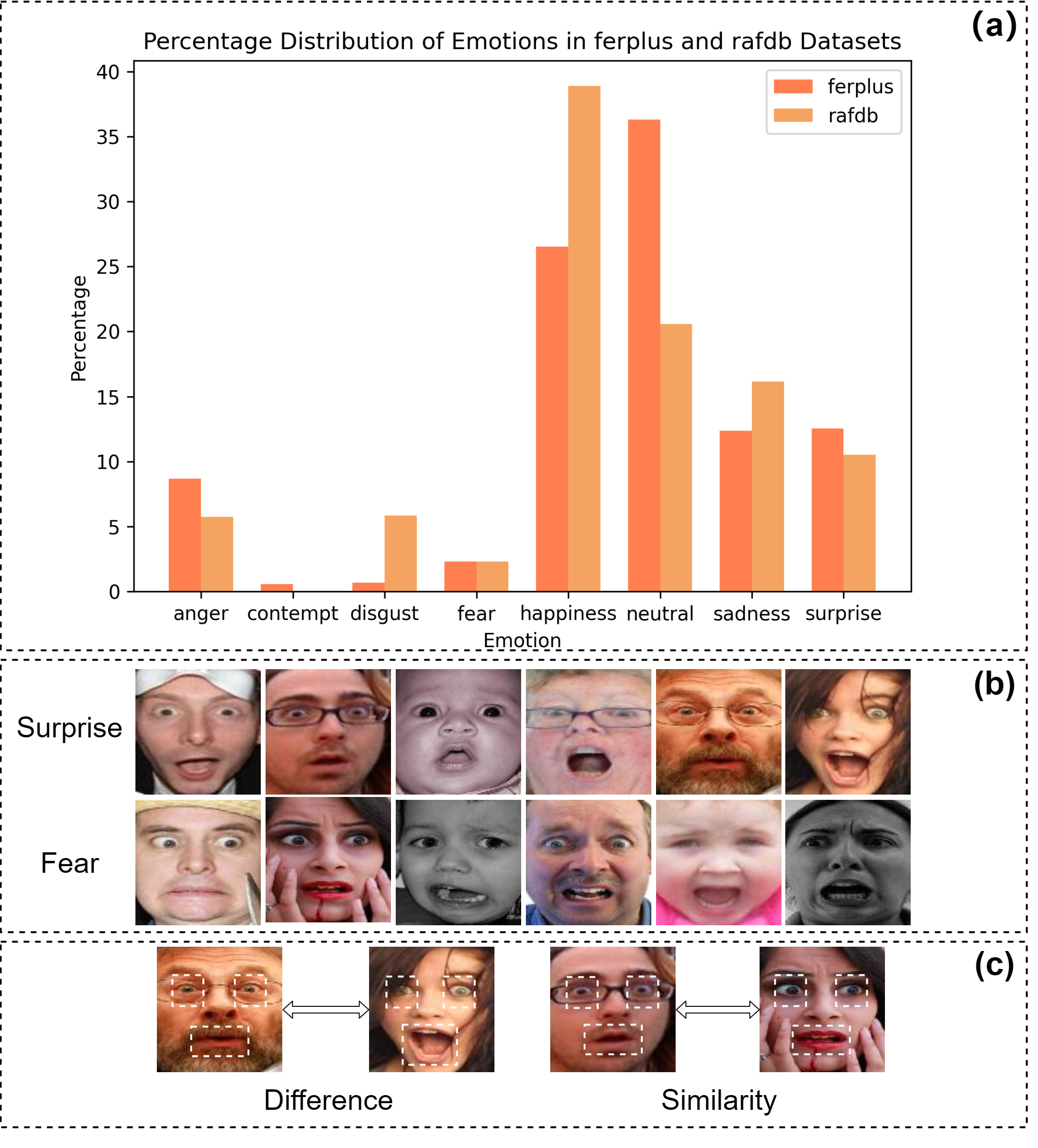}
\caption{This article attempts to address the variability between the expressions of different subjects in the field of FER. (a)Uneven distribution of data in public datasets between FERPlus and RAF-DB. (b)Similarities between expression categories. (c)Differences between different contributors in the same category.}
\label{fig-1}
\end{figure}

Facial expression recognition (FER) is one of the hot topics in the field of artificial intelligence research today. As a component of basic emotion recognition, facial expression recognition can effectively capture and analyze subtle changes in facial expressions, thereby determining a person's emotional and psychological\cite{liu2022opo} state. Accurate facial expression recognition\cite{Wang_2023_CVPR} not only provides deep insights into human emotions\cite{Liu2024}, but also lays a solid foundation for developing interactive systems with engineering ethics\cite{zhangEmotion}. However, unlike other image segmentation tasks, the challenge for Facial Expression Recognition (FER) mainly lies in the similarities between different categories of expressions\cite{Li2022TowardsSD} and the differences among different subjects within the same category\cite{Wu2023LANetLL}. For example, in Figure \ref{fig-1}B, surprise and fear are quite similar, while in \ref{fig-1}C, even though it is the same expression of surprise, there is a noticeable difference between the expressions of the elderly and the young girl. Additionally, uneven data distribution is also one of the challenges faced\cite{Zhang2023LeaveNS}. We have demonstrated the data distribution of commonly used datasets in Figure \ref{fig-1}A.

In the feature extraction stage, traditional FER methods usually use machine learning algorithms such as FERM\cite{2023SVM}, MLP\cite{2020MLP}, etc. However, these methods are difficult to overcome the detail feature extraction of facial expressions and lack robustness to facial poses or occlusions. Therefore, more and more researchers are turning their attention to deep learning models, such as convolutional neural networks and Vits models. Wang et al. \cite{2021Information} proposed the Information Reuse Attention Module, a CNN based FER method for extracting attention perception features of faces. Wang Kai et al. \cite{Wang2020SuppressingUF} proposed the SCN model, an FER model built on traditional CNN, and trained and tested it on multiple datasets. Ma et al. \cite{2021Facial} applied the Transformer model to facial expression recognition, designed the VTFF model, and designed the ASF module to fuse global and local facial information. Xue et al.\cite{APVit2022} proposed an APP and ATP module, where APP was used to select the patch with the highest information content on CNN features, and ATP was used to discard unimportant tokens in ViT. Although these works have achieved good results, they still have not effectively addressed the challenges faced by FER, namely the uneven distribution of facial expression datasets and the differences and similarities between expressions.

In order to address the challenges previously mentioned, a convolutional neural network with a ConvNeXt network as the backbone is proposed. This is known as Conv-Cut and it combines the strengths of convolutional neural networks and Transformer architectures. The aim of this is to improve the recognition accuracy of different expression categories. Specifically, the truncated ConvNeXt network is utilised as the backbone network, with the objective of reducing the number of network parameters to address the issue of the small sample problem in the FER task. In addition, inspired by the contemplation of attention mechanisms in \cite{qian2024deep}, an independent detail extraction module has been devised for the purpose of extracting fine-grained facial features. The model's capacity to attend to facial fine-grained feature regions is enhanced through the utilisation of a self-attention mechanism, thereby addressing the challenges posed by individual variability and inter-category similarity.

Overall, our contributions can be summarized as follows.

(1)The Detail Extraction Block has been designed for the purpose of extracting fine-grained features in conjunction with the self-attention mechanism. This ensures that the model pays more attention to the fine-grained feature region of the face, thereby solving the problems of individual variability and inter-category similarity.

(2) A convolutional network, Conv-cut, is proposed, based on ConvNeXt-Base. This uses truncated ConvNeXt as a feature extractor, and solves the problem of small samples in FER tasks by reducing the number of network parameters. The application of a Detail Extraction Block is intended to extract detailed features, thereby reducing the risk of overfitting due to the limited number of parameters. The experimental results demonstrate the efficacy of Conv-cut in challenging conditions.

(3) This Conv-Cut method achieved a success rate of 97.33\% on the RAF-DB dataset and 95.69\% on the FERPlus dataset, thereby demonstrating superiority over existing State-Of-The-Art methods.

\section{Related Work}

The conventional techniques for expression recognition primarily depend on manual features, categorised as low-level features, for identification. These features encompass geometric, texture, and motion characteristics.For instance, Tian et al.\cite{tian2001} have proposed a geometric feature-based expression recognition (FER) method for expression classification, which involves the identification and analysis of facial action units. In contrast, Shan et al.\cite{shan2009} have employed local binary patterns (LBP) as texture features for expression recognition. Building upon this, other researchers have extended the original technique to enhance the accuracy of expression recognition by reducing the redundancy of the LBP operator. In addition, Bartlett et al.\cite{bartlett2006} proposed an expression recognition method combining geometric and dynamic features, which achieved good results in practical applications. Alhussan et al.\cite{2023SVM} proposed an expression recognition method based on optimised support vector machines (SVMs), highlighting the key role of model optimisation and feature extraction in improving recognition performance. Despite the success of these traditional methods in the field of expression recognition, they often lack sufficient robustness in the face of complex environmental factors such as changing lighting conditions, differences in facial poses, and occlusion.

The remarkable accomplishments of deep learning in numerous domains have prompted researchers to explore its application in the domain of expression recognition. For instance, Wang et al.\cite{2021Information} have proposed an expression recognition method based on convolutional neural networks (CNNs), with a focus on information reuse and attention mechanisms. Addressing the challenge of label uncertainty in expression recognition, Wang et al.\cite{Wang2020SuppressingUF} relabelled samples with low classification confidence. Liu et al.\cite{liu2022adaptive} They then focused on extracting different fine-grained features to learn the underlying diversity and key information of facial expressions. Ma et al.\cite{2021Facial} applied the Transformer model to facial expression recognition, designed the VTFF model, and designed the ASF module to fuse global and local facial information. Xue et al.\cite{APVit2022} proposed the APP and ATP modules, where the APP is used to select the CNN feature information patches with the highest content and ATP is used to discard unimportant tokens in ViT. These works have achieved good results but still failed to effectively address the challenges faced by FER, i.e., the uneven distribution of facial expression datasets and dissimilarities between expressions.

In contrast to the aforementioned studies, the present study employs a truncated ConvNeXt network to extract facial expression features. This is achieved by reducing the number of network parameters in order to address the issue of the small sample size that is prevalent in expression recognition tasks. In addition, in order to reduce the risk of overfitting, a separate detail extraction module has been designed to extract fine-grained facial features, thereby enabling the accurate learning of the key information contained within facial expressions.

\section{Methodology}

\begin{figure}[!ht]
\centering
\includegraphics[width=0.7\textwidth]{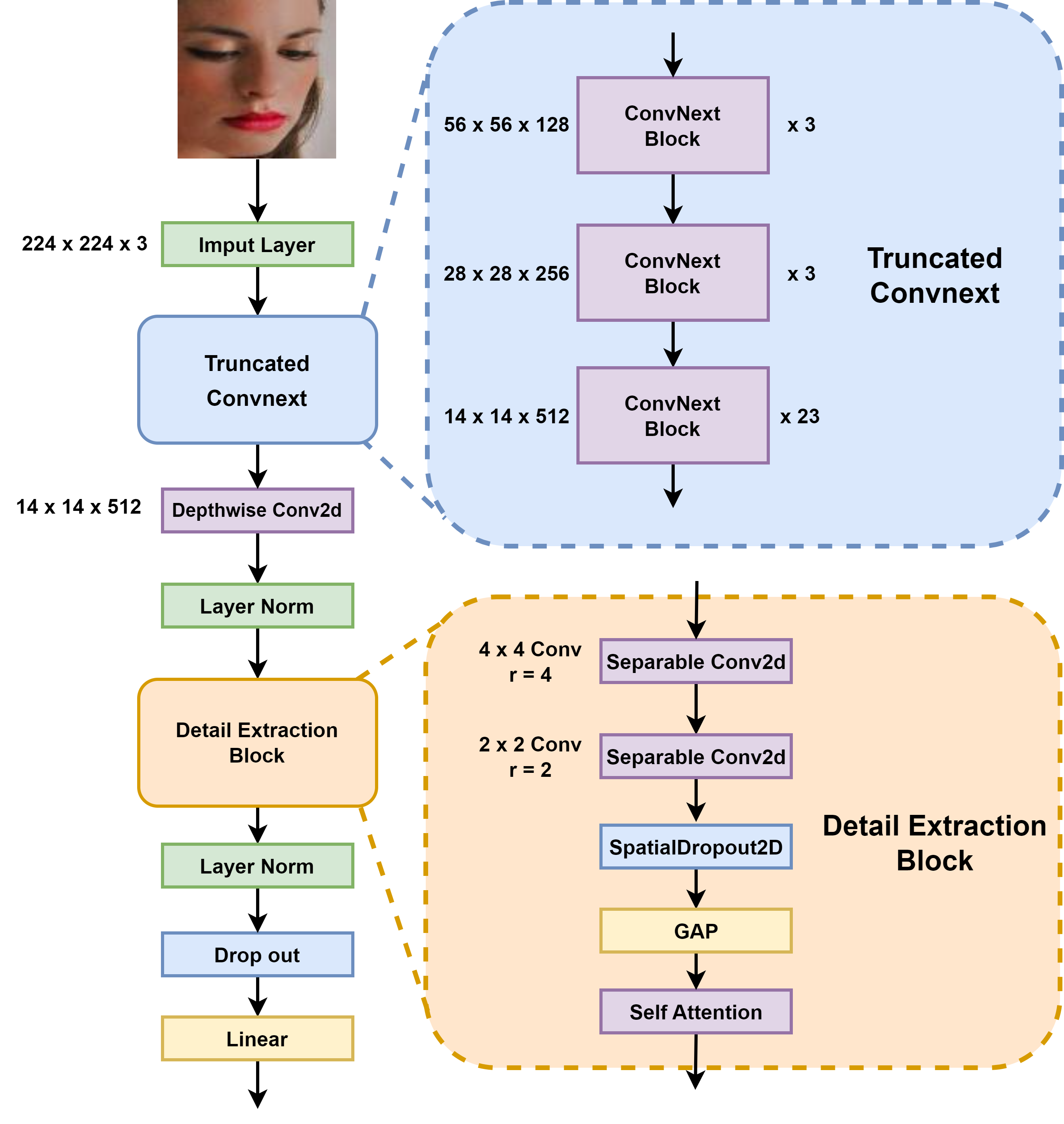}
\caption{The truncated ConvNext is utilised as the foundational framework for feature extraction, with the subsequent extraction of fine-grained features being achieved via the detail extraction module. The employment of the attention mechanism serves to enhance the model's focus on fine-grained feature regions.}
\label{fig-2}
\end{figure}

The truncated ConvNext is utilised as the fundamental framework for feature extraction, with subsequent refinement via the implementation of Detail Extraction. The overall structure of Conv-cut is illustrated in Figure \ref{fig-2}.

\textbf{Truncated Convnext.}
In comparison to other visual tasks, FER tasks are distinguished by their utilisation of small sample sizes and the presence of subtle variations in facial expressions. This characteristic poses a significant challenge in the training of even small parameter versions of the model on the available dataset. To address this challenge, a truncated network is devised, drawing upon the principles of ConvNeXt-Base. Specifically, we construct a powerful feature extraction network by truncating the third and fourth stages of ConvNeXt-Base.

It is noteworthy that ConvNeXt-Base was originally divided into four stages. In the initial two phases, the model extracts low-level features and progressively reduces the resolution of the image to one-half of the original. In the subsequent phases, the network typically extracts more high-level semantic features, with the structure of these phases primarily designed for compatibility with downstream tasks (e.g. target detection).

In order to address the characteristics of the FER task, the third and fourth stages of ConvNeXt-Base are truncated, with the feature extraction capability of the first three stages being retained and the irrelevant part of the high-level semantic extraction being removed. This approach serves to reduce the complexity and the number of model parameters, thereby enabling the model to focus on the low-level and intermediate-level features of facial expressions. This, in turn, leads to enhanced training efficiency and generalisation ability of the model when working with a limited sample size.

\textbf{Detail Extraction.}
The truncated Convnext has been demonstrated to be capable of extracting low and medium-level features of facial expressions with a reasonable degree of proficiency. However, it is observed to be deficient in its ability to extract fine-grained features. To address this limitation, a detail extraction module has been designed, consisting of a combination of convolutional network and self-attention mechanism to further extract fine-grained features of facial expressions.
As illustrated in Algorithm \ref{algo:net}, we employ two depth-separable convolutional layers to extract fine-grained features from low-level features.The first depth-separable convolutional layer possesses a convolutional kernel size of 4 X 4 and a step size of 4, which is designed to learn fine-grained features from a broader perspective and capture complete contextual relationships. The second depth-separable convolutional layer, with a convolutional kernel size of 2 X 2 and a step size of 2, is employed to further learn more detailed information. Subsequent to this, the model is enhanced through the implementation of pooling operations, with the objective of focusing on fine-grained feature regions of facial expressions, such as the corners of the eyes and mouth, in conjunction with the self-attention mechanism. Assuming the input vector is denoted by X, the query vector by Q, the key vector by K, and the value vector by V, the pooling operation and the attention operation can be expressed as follows:
\begin{equation}
x_{\text{pool}} = \frac{1}{H \times W} \sum_{i=1}^{H} \sum_{j=1}^{W} x_{i,j,:}
\end{equation}
\begin{equation}
    Attention(Q,K,V) = \mathrm{softmax}(\dfrac{QK}{\sqrt{d_{q}}})V.
\end{equation}

\begin{algorithm}[ht!]
    \caption{Pseudo code for Detail Extraction.}
\textbf{Input:} $X\in \mathbb{R}^{B \times H \times W \times C}$

\textbf{Output:} $f$: feature

$Y = \gamma \left( \frac{X - \mu}{\sqrt{\sigma^2 + \epsilon}} \right) + \beta$

$Y = SeparableConv(Y, kernelsize=4, strides=4)$

$Y = SeparableConv(Y, kernelsize=2, strides=2)$

$Y = SpatialDropout(Y, p = 0.1)$

$y_c = \frac{1}{H \times W} \sum_{i=1}^{H} \sum_{j=1}^{W} Y_{ijc}$

$Y = [y_1, y_2, ... y_c]$


$f = \text{softmax}\left( \frac{YY^T}{\sqrt{d}} \right) Y$

$ return$ $f$
    \label{algo:net}
\end{algorithm}

\section{Experiments}
\subsection{Implementation details}
A pre-trained Convnext-Base model is employed, with the weights being truncated and frozen. The hyperparameters were set as follows: batch size of 16, epoch of 50, initial learning rate of 1e-3, augmented dataset using random level-flip, sparse classification cross entropy as loss function and Adam as optimiser. All models were trained using a 16GB 4060Ti GPU.

\subsection{Datasets}
\textbf{RAF-DB}(The Real-world Affective Faces Database) \cite{RAF-DB} is a widely used in-the-wild dataset, containing over 30,000 distinct facial expression images downloaded from the internet, with both single-label and compound-label types.

\noindent \textbf{FERPlus}\cite{FERPlus} is an extension of FER2013\cite{FER2013}, collected by Google search engine, which includes 28,709 training images, 3,589 validation images, and 3,589 test images.

\subsection{Comparison with the State-of-the-art Methods}

\begin{table}[!ht]
\centering
\begin{tabular}{@{}cp{1cm}|cc}
\toprule
\multirow{2}{*}{Method} & \multirow{2}{*}{Year} & \multicolumn{2}{c}{\textbf{Acc↑(\%)}} \\
& & RAF-DB & FERPlus \\
\midrule
DMUE \cite{DUME}& 2021& 88.76&-\\
VTFF\cite{2021Facial}& 2021& 88.14&-\\
EAC\cite{Zhang2022LearnFA}& 2022& 90.35&90.05\\
POSTER\cite{Zheng2022POSTERAP}& 2023& 92.05&91.62\\
CLEF\cite{Zhang2023WeaklySupervisedTC}& 2023          & 90.09&89.74\\
APViT\cite{APVit2022}& 2023& 91.98&-\\
PAtt-Lite$^{\dag}$\cite{PAtt-Lite}& 2023& \textbf{95.05}&\textbf{93.72}\\
LNSU-Net\cite{Zhang2023LeaveNS}& 2024& 89.77&88.81\\
MGR$^{3}$Net\cite{MGR3Net}& 2024& 91.05&-\\
CEPrompt\cite{CEPrompt}& 2024& 92.43&-\\ 
\midrule
Conv-cut(Ours)&  & \textbf{97.33}(+2.28)&\textbf{95.69}(+1.97)\\
\bottomrule
\end{tabular}
\caption{Comparing accuracy of our models with the state-of-the-art methods in RAF-DB and FERPlus datasets.$^{\dag}$ is Results based on local reproduction.}
\label{tab1}
\end{table}

As illustrated in Table\ref{tab1}, the performance of our Conv-cut is demonstrated on the RAF-DB and FERPlus datasets, with a comparison to other methods over the past three years. The Conv-cut demonstrates notable efficacy in the 7-classification task, exhibiting substantial superiority over competing methods and attaining optimal outcomes. Specifically, on the RAF-DB dataset, Conv-cut demonstrates superior performance in comparison to the second-ranked PAtt-Lite method, exhibiting an enhancement of 2.28 percentage points on the ACC metric. This superiority is further substantiated in the FERPlus dataset, where an enhancement of 1.97 percentage points on the ACC metric is observed. The best results are highlighted in bold. The PAtt Lite result is obtained by replicating the original author's steps on a Nvidia 4090 GPU.

\begin{figure*}[!ht]
    \centering
    \includegraphics[width=0.7\linewidth]{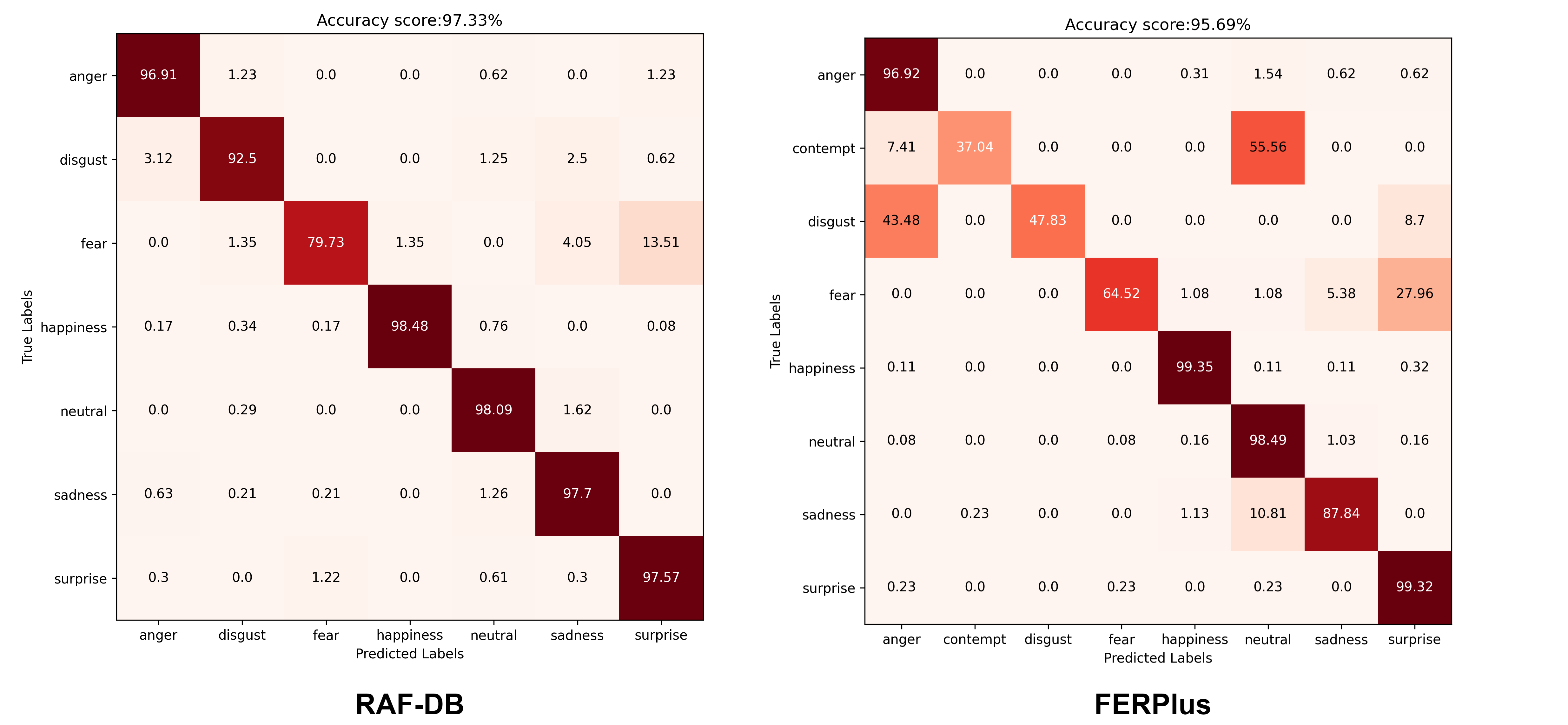}
    \caption{The confusion matrix of our proposed Conv-cut evaluated on RAF-DB and FERPlus.}
    \label{fig-3}
\end{figure*}

\begin{table}[!ht]
\centering
\small
\begin{tabular}{cccc@{}}
\toprule
 \textbf{Self-Attention}&              \textbf{Detail Extraction}& \textbf{RAF-DB↑(\%)}&                       \textbf{FERPlus↑(\%)}\\ \midrule
 $\times$& $\times$& 75.62& 75.71\\
 $\checkmark$& $\times$& 78.65(+3.03)& 78.31(+2.6)\\
 $\times$& $\checkmark$& 84.24(+5.59)& 82.46(+4.15)\\
 $\checkmark$& $\checkmark$& \textbf{97.33}(+13.09)& \textbf{95.69}(+13.23)\\\midrule
\end{tabular}
\caption{Ablation studies on the accuracy of proposed method in the RAF-DB and FERPlus datasets.}
\label{tab2}
\end{table}

\begin{table}[!ht]
\centering
\small
\begin{tabular}{l|cc@{}}
\toprule
 \textbf{Detail Extraction}& \textbf{Acc↑(\%)}& \textbf{F1↑(\%)}\\ \midrule
 One layer& 92.86& 86.84\\
                      Three layers& 95.24(+2.38)& 90.91(+4.07)\\
                      Two layers& \textbf{97.33}(+2.09)& \textbf{94.69}(+3.78)\\ \bottomrule
\end{tabular}
\caption{Comparison of different convolutional layers in the Detail Extraction module.}
\label{tab3}
\end{table}

\subsection{Ablation Studies}
\begin{figure*}[!ht]
    \centering
    \includegraphics[width=0.7\linewidth]{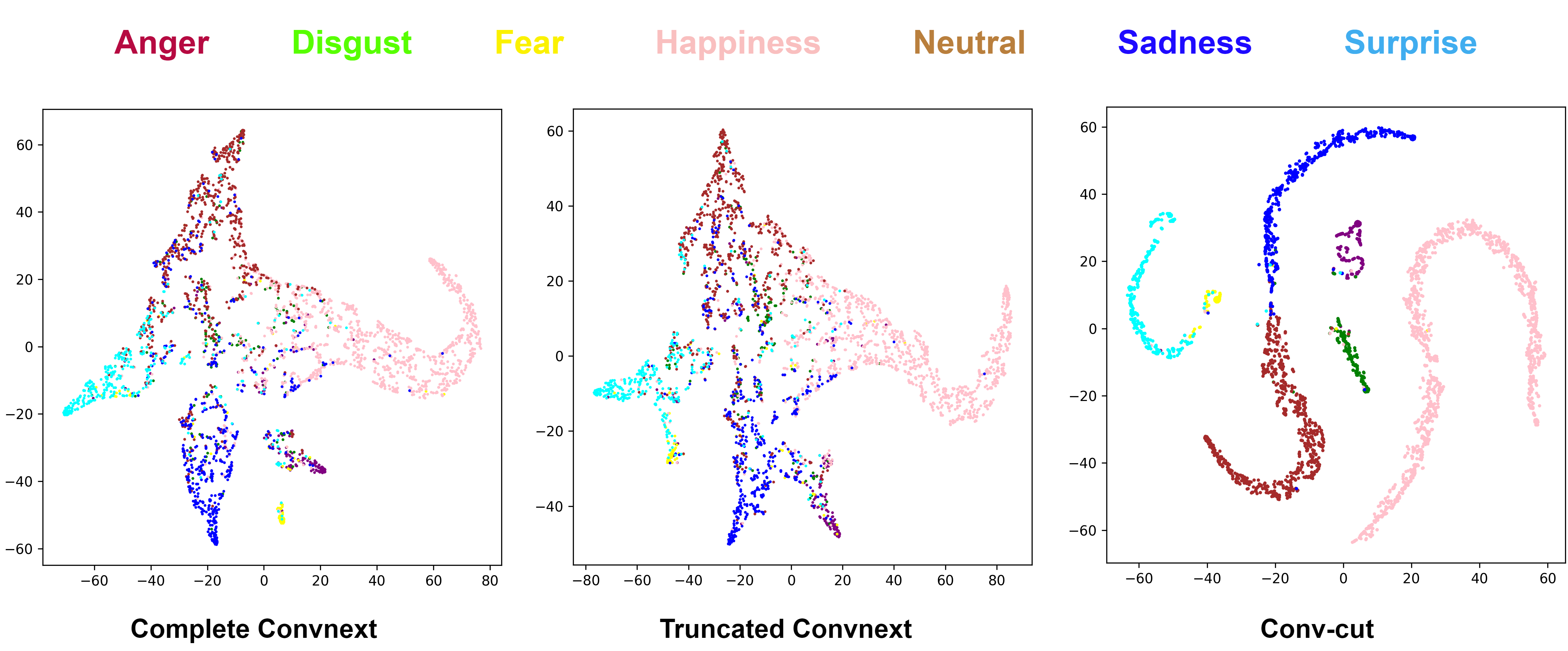}
    \caption{2D t-SNE visualization of facial expression features obtained through different models, including baseline model, truncated ConvNext model, Conv-cut. These features were extracted from the RAF-DB dataset.}
    \label{fig-4}
\end{figure*}

The full ConvNeXt Base, pre-trained on the ImageNet dataset, is utilised as the baseline model, and it is trained employing the same training method. A comparison of the detail extraction module with attention to the baseline model is then made in order to evaluate their effectiveness. The outcomes of this investigation are presented in Table\ref{tab2}. The findings reveal that employing attention alone results in a modest enhancement in model performance. The enhancement observed is approximately 3\%, as compared to the benchmark model. This finding suggests that self-attention facilitates the model's comprehension of contextual information. Conversely, the application of the Detail Extraction (DET) module without attention enhances the model's accuracy by nearly 10\%. As illustrated in Figure\ref{fig-4}, the features extracted from the baseline model are almost indistinguishable from those extracted from the truncated model. In addition, an investigation was conducted into the effect of the convolutional network layer of the detail extraction module on the model accuracy and F1 score. The experimental results are presented in Table\ref{tab3}, which indicates that the module demonstrates optimal fine-grained feature extraction capability when the number of convolutional layers is two.

\begin{figure*}[!ht]
    \centering
    \includegraphics[width=0.7\linewidth]{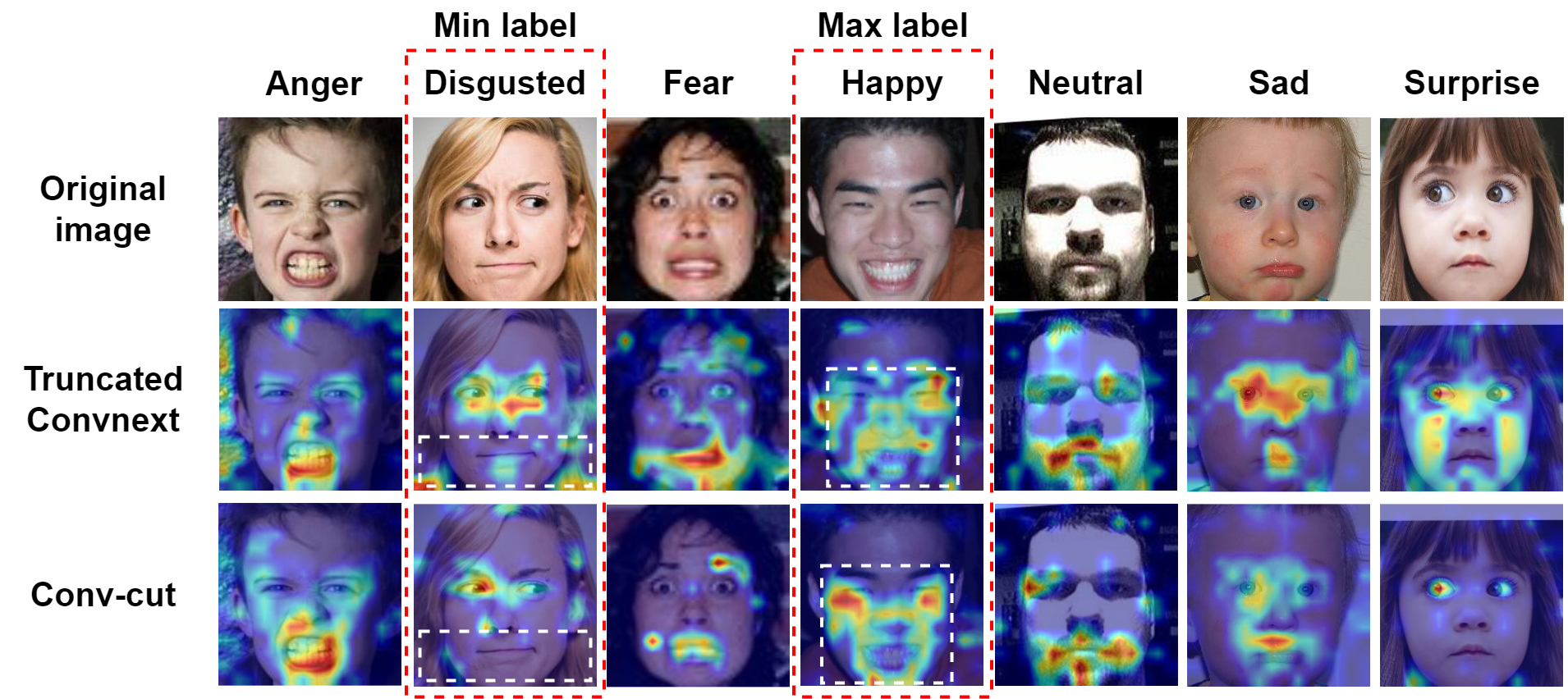}
    \caption{Attention visualization for images from RAF-DB dataset. The min label and max label represent the average minimum and average maximum number of categories, respectively.}
    \label{fig-5}
\end{figure*}
\subsection{Visualization}
\textbf{Attention Visualization. }In order to further investigate the effectiveness of Conv-cut, Grad-CAM was utilised to visualise randomly sampled attention maps from the RAF-DB dataset.As illustrated in Figure\ref{fig-5}, the initial row depicts the original image from left to right, with 'angry', 'disgusted', 'fearful', 'happy', 'neutral', 'sad' and 'surprised' from left to right. A comparison of the second and third rows reveals that the truncated Convnext focuses predominantly on expression-related regions, yet still focuses on redundant features (e.g., the attention to hair in 'disgust').Con-cut eliminates redundant features and accurately focuses on expression-related regions (e.g., the corners of the eyes in 'neutral' and the lips in 'sad'). In the case of category imbalance, the average minimum number of samples is shown to be as good as the average maximum number of samples for focusing on.

\noindent \textbf{Feature Visualization .}We use T-SNE \cite{tsne} to visualize the distribution of facial expression features extracted from our benchmark model and Conv-cut. We also visualize the confusion matrix of our proposed CFER model evaluated on the RAF-DB and FERPlus datasets to analyze the results. In Figure \ref{fig-3}, we observe that the model evaluated on RAF-DB had a lower prediction accuracy for negative emotions such as fear, with only 79.73\%. This is because the proportion of such negative emotions in the RAF-DB dataset is relatively low, with fear accounting for only 2.28\%, resulting in poor performance of the model for such emotions. In addition, we also observe that the models evaluated on FERPlus had lower accuracy in predicting contemper, disgust, and fear. The models frequently incorrectly predict contemper as neutral because contemper and disgust have a relatively low proportion in FERPlus, with contemper accounting for 0.58\%, disgust accounting for 0.67\%, and fear accounting for 2.29\%. Therefore, the models are more inclined to predict these negative emotions as other emotions.

\section{Conclusion}
In this paper, we propose a convolutional neural network with ConvNeXt as the backbone, which still performs well in challenging situations. We observe through experiments that the features obtained by the complete ConvNeXt model and the truncated ConvNeXt model are basically the same, and the detailed features can be better extracted by building the Detail Extraction module, which provides a new idea for future work.

However, due to the uneven distribution of FERPlus data, our model performs poorly in predicting negative expressions on FERPlus and is also less stable in training compared to RAF-DB. Our follow-up work will focus on solving this problem.

\begin{credits}
\subsubsection{\ackname} This work supported by Beijing Key Laboratory of Behavior and Mental Health, Peking University.

\subsubsection{\discintname}
The authors have no competing interests to declare that are relevant to the content of this article.
\end{credits}
%
%
%
\bibliographystyle{splncs04}
\bibliography{paper}

\end{document}